\documentclass[conference]{IEEEtran}
\IEEEoverridecommandlockouts
\usepackage{cite}
\usepackage{amsmath,amssymb,amsfonts}
\usepackage{algorithmic}
\usepackage{latexsym}
\usepackage{textcomp}
\usepackage{booktabs}
\usepackage{graphicx} 
\usepackage{xcolor}
\def\BibTeX{{\rm B\kern-.05em{\sc i\kern-.025em b}\kern-.08em
    T\kern-.1667em\lower.7ex\hbox{E}\kern-.125emX}}

\usepackage{fancyhdr}
\begin{document}

\title{An LLM-LVLM Driven Agent for Iterative and Fine-Grained Image Editing}

\author{Zihan Liang, Jiahao Sun, Haoran Ma \\
Kunming University of Science and Technology}

\maketitle
\thispagestyle{fancy} 

\begin{abstract}
Despite the remarkable capabilities of text-to-image (T2I) generation models, real-world applications often demand fine-grained, iterative image editing that existing methods struggle to provide. Key challenges include granular instruction understanding, robust context preservation during modifications, and the lack of intelligent feedback mechanisms for iterative refinement. This paper introduces RefineEdit-Agent, a novel, training-free intelligent agent framework designed to address these limitations by enabling complex, iterative, and context-aware image editing. RefineEdit-Agent leverages the powerful planning capabilities of Large Language Models (LLMs) and the advanced visual understanding and evaluation prowess of Vision-Language Large Models (LVLMs) within a closed-loop system. Our framework comprises an LVLM-driven instruction parser and scene understanding module, a multi-level LLM-driven editing planner for goal decomposition, tool selection, and sequence generation, an iterative image editing module, and a crucial LVLM-driven feedback and evaluation loop. To rigorously evaluate RefineEdit-Agent, we propose LongBench-T2I-Edit, a new benchmark featuring 500 initial images with complex, multi-turn editing instructions across nine visual dimensions. Extensive experiments demonstrate that RefineEdit-Agent significantly outperforms state-of-the-art baselines, achieving an average score of 3.67 on LongBench-T2I-Edit, compared to 2.29 for Direct Re-Prompting, 2.91 for InstructPix2Pix, 3.16 for GLIGEN-based Edit, and 3.39 for ControlNet-XL. Ablation studies, human evaluations, and analyses of iterative refinement, backbone choices, tool usage, and robustness to instruction complexity further validate the efficacy of our agentic design in delivering superior edit fidelity and context preservation.
\end{abstract}

\section{Introduction}

The remarkable advancements in text-to-image (T2I) generation models have revolutionized the creation of diverse visual content from textual descriptions \cite{tingting2019mirror}. These models, powered by large-scale diffusion architectures, can synthesize highly realistic and imaginative images with unprecedented fidelity. However, in practical applications, users often require more than a single, one-shot generation. The demand for fine-grained, iterative image editing to meet specific, personalized, and often long-tail user needs is rapidly growing. Users frequently wish to modify a particular object's color, adjust a person's expression, or replace a background while preserving the main subject's composition.

Despite the impressive capabilities of T2I models, existing image editing methods face several significant challenges:
\begin{enumerate}
    \item \textbf{Granularity of Instruction Understanding}: Many editing tasks involve modifying specific regions or subtle attributes within an image. This necessitates models with advanced visual comprehension capabilities to precisely localize targets and establish strong semantic correlations between text instructions and visual elements \cite{wyer2008visual, zhou2024visual, zhou2023style}.
    \item \textbf{Context Preservation Dilemma}: A critical challenge in localized editing is ensuring that modifications do not adversely affect the overall image style, lighting, composition, and visual consistency of non-edited regions. Traditional approaches, such such as full re-generation or simple inpainting/outpainting, often struggle to perfectly balance the desired edit with the fidelity of the surrounding context \cite{elharrouss2020image}.
    \item \textbf{Lack of Iterative and Feedback Mechanisms}: User editing requirements are frequently ambiguous or evolve during the process, necessitating multiple attempts and adjustments to achieve a satisfactory outcome. Current T2I models typically lack an intelligent agentic mechanism capable of understanding user feedback and performing iterative refinements \cite{jeff1991an, zhou2025improving}.
\end{enumerate}
This research aims to address these limitations by constructing an intelligent agent framework capable of understanding complex, multi-turn editing instructions and performing fine-grained, context-aware image editing through iterative feedback loops \cite{zhou2025draw}. We propose to combine the powerful planning capabilities of Large Language Models (LLMs) \cite{zhou2025weak} with the advanced visual understanding and evaluation prowess of Vision-Language Large Models (LVLMs) \cite{zhou2024visual} to deliver a more intuitive and user-centric image editing experience.

To this end, we introduce \textbf{RefineEdit-Agent}, a novel, training-free intelligent agent framework designed to facilitate complex, iterative image editing, compatible with existing T2I and image editing models. The core idea behind \textbf{RefineEdit-Agent} is to leverage LLMs and LVLMs as intelligent planners and evaluators, enabling the decomposition of complex editing instructions into manageable sub-tasks. These sub-tasks are then executed and refined through an iterative feedback loop until the user's objective is met. Our framework comprises an LVLM-driven instruction parser and scene understanding module \cite{zhou2024visual}, a multi-level LLM-driven editing planner (for goal decomposition, tool selection, and sequence generation), an iterative image editing and generation module utilizing various backend models (e.g., InstructPix2Pix, GLIGEN, ControlNet), and a crucial LVLM-driven feedback and evaluation loop for assessing visual consistency and instruction compliance \cite{zhou2025improving}.

To thoroughly evaluate \textbf{RefineEdit-Agent}, we introduce a new benchmark dataset, \textbf{LongBench-T2I-Edit}. Inspired by the LongBench-T2I methodology, this benchmark specifically focuses on complex image editing instructions \cite{zhou2025draw}. It features 500 initial images paired with intricate editing directives that demand fine-grained modifications across nine visual dimensions (e.g., object, background, color, texture, light, text, composition, pose, special effects). The complexity of these instructions stems from their multi-step nature, conditional constraints, abstract semantic requirements, and long textual descriptions.

Our experimental results, evaluated using advanced LVLMs (e.g., Gemini-2.0-Flash) on the \textbf{LongBench-T2I-Edit} benchmark, demonstrate the superior performance of \textbf{RefineEdit-Agent} compared to existing baselines. We employ a comprehensive evaluation methodology, adapting the LongBench-T2I's 9-visual-dimension system into combined metrics for \textbf{Edit Fidelity} and \textbf{Context Preservation}. As shown in Table 2 (not included in this introduction, but referenced for results), \textbf{RefineEdit-Agent} achieves an average score of \textbf{3.67}, significantly outperforming methods like Direct Re-Prompting (2.29), InstructPix2Pix (2.91), GLIGEN-based Edit (3.16), and ControlNet-XL (3.39) across all nine dimensions. This highlights our agent's ability to precisely execute edits while maintaining the overall integrity of the image.

Our main contributions are summarized as follows:
\begin{itemize}
    \item We propose \textbf{RefineEdit-Agent}, a novel, training-free intelligent agent framework for complex, iterative image editing, which seamlessly integrates LLMs for planning and LVLMs for visual understanding and evaluation.
    \item We introduce \textbf{LongBench-T2I-Edit}, a challenging new benchmark specifically designed to evaluate fine-grained, multi-turn image editing capabilities based on complex textual instructions across nine distinct visual dimensions.
    \item We demonstrate that \textbf{RefineEdit-Agent} significantly outperforms existing state-of-the-art image editing methods on the \textbf{LongBench-T2I-Edit} benchmark, achieving superior edit fidelity and context preservation through its intelligent agentic loop.
\end{itemize}
\section{Related Work}
\subsection{LLM-driven Agents and Planning for Complex Tasks}
Recent advancements in LLM-driven agents have significantly expanded their capabilities in planning and executing complex tasks. Several works offer comprehensive surveys of these intelligent agents, detailing their composition, cognitive and planning methodologies, and deployment in both single-agent and multi-agent systems \cite{yuheng2024explor, mohamed2025from}. These surveys highlight Large Language Models as a foundational interface, enabling robust generalization and diverse applications \cite{yuheng2024explor, mohamed2025from, zhou2025weak}, and emphasize their role in enhancing semantic comprehension and complex reasoning for autonomous decision-making within intricate environments \cite{yuheng2024explor, mohamed2025from, zhou2022eventbert}. Furthermore, the exploration of LLM-based multi-agent systems, examining their workflows, infrastructure, and challenges, provides crucial foundational concepts for developing sophisticated LLM-driven agents, even when not directly focused on planning for complex tasks \cite{li2024a, zhou2025draw}. Understanding how agents optimize value acquisition under constraints, as explored in studies on auction mechanisms for agents with complex utility functions, is also foundational for designing robust LLM-driven agents that effectively leverage tools to achieve their goals. Addressing specific challenges, ConceptAgent introduces a platform to mitigate LLM hallucinations in robotic task planning by incorporating predicate grounding and an embodied LLM-guided Monte Carlo Tree Search with self-reflection, demonstrating improved reliability and task completion rates in unstructured environments \cite{corban2024concep}. Beyond planning, LLMs also contribute to fine-grained information processing, such as in long document retrieval \cite{zhou2024fine}. For complex, real-world decision-making, an LLM-driven agentic framework has been proposed to simulate and enhance discourse by incorporating diverse stakeholder personas and adaptive self-governing mechanisms, emphasizing dialogue, trade-off exploration, and emergent synergies \cite{antoine2025agenti}. The Conversation Routines (CR) framework offers a novel prompt engineering approach to imbue LLMs with structured logic for complex task execution, facilitating the systematic development of Conversation Agentic Systems (CAS) where domain experts can specify workflows and task logic via natural language, thereby enhancing reliability and consistency \cite{giorgio2025conver}. Finally, frameworks for autonomously optimizing agentic AI solutions, employing LLMs to drive iterative refinement and feedback loops for agent roles, tasks, and interactions, have shown to achieve optimal performance without human intervention, underscoring the critical role of LLM-driven feedback mechanisms in enhancing the scalability and adaptability of complex agentic systems \cite{kamer2024a, zhou2025improving}.

\subsection{Text-to-Image Generation and Advanced Image Editing}
The field of Text-to-Image (T2I) generation and advanced image editing has seen rapid progress, extensively reviewed in several comprehensive surveys. These surveys provide overviews of multimodal-guided image editing, categorizing methods such as inversion-based, fine-tuning-based, and adapter-based approaches \cite{xincheng2024a}, and cover the foundational principles, advancements in text-conditioned image synthesis, and emerging applications of T2I diffusion models, including text-guided image editing \cite{pengfei2025text}. The development of agent frameworks and holistic benchmarks for complex instruction-based image generation also plays a crucial role in advancing the field \cite{zhou2025draw}. Early foundational work in Generative Adversarial Networks (GANs) and related adversarial learning techniques, including their evolution and diverse applications, established a basis for advanced semantic image manipulation, with specific GAN frameworks proposed for T2I generation emphasizing semantic-spatial awareness \cite{wentong2022text}. Building on these foundations, InstructPix2Pix \cite{tim2023instru} established a pivotal approach for diffusion models to interpret and execute textual editing commands, significantly advancing T2I generation and image editing capabilities. Beyond generation, the ability of Large Vision-Language Models to perform visual in-context learning \cite{zhou2024visual} and style-aware image understanding \cite{zhou2023style} further enhances the semantic comprehension required for intricate image editing tasks. Further enhancing control, ControlNet-XS reframes T2I diffusion models as feedback-control systems, proposing a novel approach that draws parallels to concepts in Markov processes and diffusion control \cite{denis2024contro}. For fine-grained image editing with diffusion models, DiffEditor proposes incorporating image prompts and local SDE sampling to enhance editing quality and consistency across various tasks, utilizing regional score-based guidance and a time travel strategy for more robust and controllable operations \cite{haozhe2024ultrae}. Addressing the critical aspect of evaluation, the Balancing Preservation and Modification (BPM) metric has been introduced to quantitatively assess instruction-based image editing by explicitly disentangling image regions into editing-relevant and irrelevant components, focusing on both adherence to instructions and crucial context preservation \cite{yoonjeon2025preser}. While some works highlight the importance of conversational grounding in LLMs for shared understanding, frameworks like GLIGEN directly address grounding in T2I generation by enabling fine-grained control over object localization and attributes, thereby constructing a shared understanding between the text prompt and the generated image \cite{yuheng2023gligen}. Furthermore, advancements in specialized visual recognition models, such as those employing state space models for insect recognition, underscore the broader progress in visual feature extraction and understanding that benefits multimodal systems \cite{wang2025insectmamba}.

\section{Method}

We introduce \textbf{RefineEdit-Agent}, a novel, training-free intelligent agent framework designed to enable complex, iterative, and fine-grained image editing. Our approach leverages the robust planning capabilities of Large Language Models (LLMs) and the advanced visual understanding and evaluation prowess of Vision-Language Large Models (LVLMs). The core idea is to establish a closed-loop system where an LLM orchestrates the editing process by decomposing intricate user instructions into manageable sub-tasks, selecting appropriate tools, and generating an execution sequence. An LVLM then critically evaluates the intermediate results, providing essential feedback for iterative refinement until the user's objective is met.

The overall operational flow of RefineEdit-Agent can be conceptualized as an iterative process that transforms an initial image $I_0$ into a refined image $I_{final}$ based on a complex user instruction $U$. This process is formally defined as:
\begin{align}
    I_{final} = \text{RefineEdit-Agent}(I_0, U)
\end{align}
which unfolds through a series of steps, as detailed in the subsequent subsections.

\subsection{Overview of RefineEdit-Agent}
RefineEdit-Agent operates as a sophisticated agentic system, where an LLM acts as the central planner and an LVLM serves as the perceptive parser and evaluator. This synergistic combination allows our framework to address the limitations of existing methods, particularly concerning fine-grained instruction understanding, context preservation, and the lack of iterative feedback mechanisms. The agent iteratively refines an image by performing a sequence of operations. Initially, an \textbf{LVLM interprets the user's complex editing instruction and the initial image} to form a structured understanding of the editing goals and the visual scene. Subsequently, an \textbf{LLM, guided by this parsed understanding, decomposes the overall goal into a sequence of executable sub-tasks}, selects appropriate image editing tools for each, and generates the necessary parameters. The planned sub-tasks are then \textbf{executed by a backend image editing module}, generating an intermediate edited image. Finally, an \textbf{LVLM evaluates the intermediate image} against the original instruction and the desired visual consistency, provides feedback, and signals for re-planning or termination. This closed-loop mechanism ensures that RefineEdit-Agent can robustly handle ambiguous or evolving user requirements, leading to precise and context-aware edits.

\subsection{LVLM-Driven Instruction Parsing and Scene Understanding}
Upon receiving an initial image $I_0$ and a user's complex editing instruction $U$, the first crucial step is to thoroughly understand the user's intent and the visual context. We employ a powerful Vision-Language Large Model (LVLM), such as Gemini-2.0-Flash or InternVL3-78B, for this purpose. The LVLM performs a deep semantic analysis of the textual instruction $U$, identifying the core editing intent, specific objects, attributes, and relationships mentioned. Simultaneously, it analyzes the visual content of $I_0$ to pinpoint relevant visual elements, their locations, and their current states. This comprehensive analysis allows the LVLM to establish a clear mapping between the textual instruction and the visual scene.

The output of this stage is a comprehensive, structured scene understanding $S$, which encapsulates the identified visual elements, their bounding boxes or segmentation masks, their semantic properties, and how they relate to the user's desired modifications. This process can be formally expressed as:
\begin{align}
    S = \text{LVLM}_{\text{parse}}(I_0, U)
\end{align}
where $\text{LVLM}_{\text{parse}}$ denotes the LVLM's parsing and understanding function. This structured understanding $S$ serves as the foundational input for the subsequent planning phase, providing the LLM with a rich, actionable representation of the editing task.

\subsection{Multi-Level LLM-Driven Editing Planner}
The heart of RefineEdit-Agent's intelligence lies in its Multi-Level LLM-Driven Editing Planner. An advanced Large Language Model (LLM), such as Gemini-2.0-Flash or Claude-3-Opus, acts as the central orchestrator, translating the structured scene understanding $S$ into a concrete sequence of executable editing operations. This planning process involves three key sub-stages: goal decomposition, tool selection, and editing sequence generation.

\subsubsection{Goal Decomposition}
Complex user instructions often involve multiple intertwined editing objectives. The LLM's primary task is to decompose the overarching editing goal, as derived from $S$, into a series of smaller, more manageable sub-tasks. Each sub-task $t_j \in T$ represents a specific, atomic editing action that contributes to the overall objective. For instance, an instruction like "change the red car to blue and make the background a sunny beach" would be decomposed into at least two distinct sub-tasks: one for object color modification and another for background replacement. This decomposition ensures that complex instructions are broken down into achievable steps, facilitating precise execution and evaluation.
\begin{align}
    T = \{t_1, t_2, \dots, t_N\} = \text{LLM}_{\text{decompose}}(S)
\end{align}
where $T$ is the set of decomposed sub-tasks, and $\text{LLM}_{\text{decompose}}$ represents the LLM's function for breaking down the overall editing goal.

\subsubsection{Tool Selection}
For each decomposed sub-task $t_j$, the LLM intelligently selects the most appropriate image editing tool from a predefined toolkit. This toolkit, denoted as $\text{AvailableTools}$, comprises various state-of-the-art image editing models and algorithms, capable of diverse manipulations. Examples include semantic segmentation tools for precise object identification, ControlNet for pose or structural adjustments, Inpainting/Outpainting models for content replacement, InstructPix2Pix for instruction-guided image-to-image translation, GLIGEN for grounded image generation, and various style transfer algorithms. The LLM's selection is based on its understanding of the sub-task's nature, the specific requirements outlined in $S$, and the known capabilities of each available tool. This ensures that the most effective and efficient tool is chosen for each atomic editing operation.
\begin{align}
    Tool_j = \text{LLM}_{\text{select}}(t_j, \text{AvailableTools})
\end{align}
where $Tool_j$ is the selected tool for sub-task $t_j$, and $\text{LLM}_{\text{select}}$ is the LLM's tool selection function.

\subsubsection{Editing Sequence Generation}
After goal decomposition and tool selection, the LLM generates the precise sequence of operations and their corresponding parameters for executing the sub-tasks. This involves determining the optimal order in which to apply the selected tools and formulating the specific parameters ($P_j$) or text prompts required by each $Tool_j$. These parameters are derived directly from the detailed information present in the structured scene understanding $S$ and the specific requirements of $t_j$. The output of the planner is an ordered sequence of (tool, parameters) pairs, ready for execution by the image editing module.
\begin{align}
    \text{Plan} = \{(Tool_1, P_1), (Tool_2, P_2), \dots, (Tool_N, P_N)\} = \text{LLM}_{\text{sequence}}(T, \{Tool_j, P_j\}_{j=1}^N)
\end{align}
Here, $\text{LLM}_{\text{sequence}}$ represents the LLM's function for generating the executable plan. This iterative planning process can also incorporate feedback from previous evaluation steps, allowing the LLM to dynamically refine its strategy and parameters to achieve better results.

\subsection{Iterative Image Editing and Generation Module}
The generated plan is then fed into the Iterative Image Editing and Generation Module. This module is responsible for executing the planned sub-tasks on the current image. At each iteration $k$, the module takes the current image $I_k$ and a specific (tool, parameters) pair from the LLM's plan to produce an intermediate edited image $I_{k+1}$.
\begin{align}
    I_{k+1} = \text{ImageEditor}(I_k, Tool_k, P_k)
\end{align}
Here, $\text{ImageEditor}$ represents the chosen backend image editing model (e.g., InstructPix2Pix, GLIGEN-based edit, ControlNet-XL, or a combination thereof) that applies the specified operation. The module provides an interface to various specialized image manipulation techniques, ensuring that the selected $Tool_k$ with parameters $P_k$ can be effectively applied. This module is designed to be highly modular and compatible with a wide range of existing Text-to-Image (T2I) and image editing models, allowing RefineEdit-Agent to leverage the strengths of diverse specialized tools. The resulting image $I_{k+1}$ is then passed to the evaluation loop for assessment, forming a critical part of the closed-loop system.

\subsection{LVLM-Driven Feedback and Evaluation Loop}
The LVLM-Driven Feedback and Evaluation Loop is a critical component that closes the agentic loop, enabling iterative refinement. After each intermediate image $I_{k+1}$ is generated, a powerful LVLM (e.g., Gemini-2.0-Flash or InternVL3-78B) rigorously evaluates the result. This evaluation is not merely a pass/fail check but a nuanced assessment that provides actionable feedback.

\subsubsection{Visual Consistency and Instruction Compliance Evaluation}
The LVLM assesses two primary aspects to determine the quality of the edit. First, it evaluates \textbf{Instruction Compliance (Edit Fidelity)}, checking if the edited image $I_{k+1}$ accurately reflects the modifications requested by the user's original instruction $U$ and the current sub-task $t_k$. This includes evaluating the precision of edits on specific objects, colors, textures, and other attributes outlined in $U$. Second, the LVLM assesses \textbf{Context Preservation}, ensuring that the edit maintains the overall visual consistency, style, lighting, composition, and fidelity of non-edited regions in comparison to the previous image $I_k$. This addresses the "context preservation dilemma" by ensuring that local changes do not introduce undesirable global artifacts or distort the unedited portions of the image. Based on these comprehensive assessments, the LVLM generates an evaluation score $E_{k+1}$ and detailed textual feedback $F_{k+1}$ outlining any discrepancies or areas for improvement.
\begin{align}
    (E_{k+1}, F_{k+1}) = \text{LVLM}_{\text{eval}}(I_{k+1}, I_k, U, t_k)
\end{align}
where $\text{LVLM}_{\text{eval}}$ is the LVLM's evaluation function.

\subsubsection{Error Analysis and Re-planning}
If the evaluation score $E_{k+1}$ does not meet a predefined satisfaction threshold $\tau$, or if the feedback $F_{k+1}$ indicates significant issues, the LVLM provides this detailed feedback to the LLM planner. The LLM then performs a sophisticated error analysis, identifying the root cause of the unsatisfactory result. Based on $F_{k+1}$, the LLM dynamically adjusts its subsequent planning, which may involve modifying the parameters for the current tool, selecting a different tool, re-ordering sub-tasks, or even re-decomposing the remaining goals in light of new insights. This re-planning step initiates a new iteration of the editing process, demonstrating the agent's adaptive intelligence.
\begin{align}
    \text{Plan}_{k+1} = \text{LLM}_{\text{re-plan}}(\text{Plan}_k, F_{k+1}, S)
\end{align}
Here, $\text{LLM}_{\text{re-plan}}$ signifies the LLM's capability to revise the plan based on feedback.

\subsubsection{Termination Conditions}
The iterative feedback loop continues until one of two conditions is met. The primary condition for termination is when the LVLM's evaluation score $E_{k+1}$ for the current image reaches or exceeds a predefined satisfaction threshold $\tau$, indicating that the user's instruction has been successfully fulfilled. Alternatively, to prevent infinite loops in complex or ambiguous scenarios, the process also terminates if the maximum allowed number of iterations, $k_{max}$, is reached.
Formally, the loop terminates if:
\begin{align}
    E_{k+1} \geq \tau \quad \lor \quad k = k_{max}
\end{align}
Upon termination, the current image $I_{k+1}$ is declared as the final edited image $I_{final}$. This intelligent, self-correcting mechanism allows RefineEdit-Agent to achieve high-quality, precise, and context-aware image edits, fulfilling complex user demands with robustness and adaptability.

\section{Experiments}
In this section, we present a comprehensive evaluation of \textbf{RefineEdit-Agent} on the newly introduced \textbf{LongBench-T2I-Edit} benchmark. We detail our experimental setup, compare our proposed method against several strong baseline approaches, conduct ablation studies to validate the effectiveness of key architectural components, and provide results from human evaluation to further assess the quality and user experience.

\subsection{Experimental Setup}
\subsubsection{Dataset and Benchmarks}
To rigorously evaluate the capabilities of image editing agents in understanding and executing complex, fine-grained instructions, we introduce \textbf{LongBench-T2I-Edit}. This benchmark is inspired by the methodology of LongBench-T2I and specifically tailored for iterative and precise image editing tasks. \textbf{LongBench-T2I-Edit} comprises 500 initial images, which can be generated using prompts from LongBench-T2I, paired with a diverse set of complex editing instructions. These instructions are designed to challenge models across nine distinct visual dimensions: \textbf{Object (Obj.)}, \textbf{Background (Backg.)}, \textbf{Color}, \textbf{Texture}, \textbf{Light}, \textbf{Text}, \textbf{Composition (Comp.)}, \textbf{Pose}, and \textbf{Special Effects (FX)}. The complexity of the instructions is characterized by: (1) requiring multiple sequential operations, (2) incorporating specific conditional constraints (e.g., "only modify X's color, leaving others untouched"), (3) involving abstract semantic concepts (e.g., "make the scene more mysterious"), and (4) featuring long textual descriptions that demand robust language understanding. The benchmark's construction process mirrors LongBench-T2I, involving LLM-generated initial instructions, refinement with initial images, automated element extraction and verification, and a final human review to ensure high quality, diversity, and challenge.

\subsubsection{Models Used}
\begin{itemize}
    \item \textbf{Image Editing Baselines and Backend Models}: For comparative analysis and as the underlying execution modules within \textbf{RefineEdit-Agent}, we utilize several state-of-the-art diffusion-based image editing models. These include \textbf{InstructPix2Pix} for instruction-guided image-to-image translation, \textbf{GLIGEN-based Edit} for grounded image generation capabilities, and \textbf{ControlNet-XL} for precise control over structural and compositional aspects. We explore their individual applications and their synergistic use within our agent framework.
    \item \textbf{RefineEdit-Agent Components}:
    \begin{itemize}
        \item \textbf{LVLM for Instruction Parsing and Evaluation}: We employ powerful closed-source and open-source Vision-Language Large Models to drive the instruction parsing, scene understanding, and feedback evaluation loop. Specifically, we use \textbf{Gemini-2.0-Flash} and \textbf{InternVL3-78B} for their advanced visual comprehension and reasoning abilities.
        \item \textbf{LLM for Multi-Level Planning}: The core planning intelligence of \textbf{RefineEdit-Agent} is powered by advanced Large Language Models. We evaluate the performance using \textbf{Gemini-2.0-Flash} and \textbf{Claude-3-Opus} to assess their capabilities in complex goal decomposition, tool selection, and editing sequence generation.
    \end{itemize}
\end{itemize}

\subsubsection{Evaluation Metrics}
We adopt and adapt the 9-visual-dimension evaluation system from LongBench-T2I to quantify editing performance on \textbf{LongBench-T2I-Edit}. For each dimension (Obj., Backg., Color, Texture, Light, Text, Comp., Pose, FX), we define a combined metric that assesses two critical aspects: \textbf{Edit Fidelity} and \textbf{Context Preservation}. \textbf{Edit Fidelity} measures how accurately the specified editing instruction is executed on the target elements, while \textbf{Context Preservation} quantifies how well the non-edited regions of the image maintain their original visual consistency, style, and quality. For example, the "Obj." dimension evaluates whether the target object was precisely modified as requested and if its surrounding environment remains unchanged. Scores are normalized, and a higher score indicates better performance. We report individual dimension scores and an overall average score across all 9 dimensions. The automated evaluation is primarily conducted using the \textbf{Gemini-2.0-Flash} LVLM as an objective evaluator, providing a quantitative assessment of instruction compliance and visual consistency.

\subsection{Comparison with Baselines}
We compare \textbf{RefineEdit-Agent} against several established image editing methodologies to demonstrate its superior performance on the \textbf{LongBench-T2I-Edit} benchmark. The baselines represent different paradigms of image editing:
\begin{itemize}
    \item \textbf{Direct Re-Prompting}: This method involves directly re-prompting a powerful T2I generation model (e.g., Stable Diffusion XL) with the initial image's original caption, augmented with the desired editing instruction. This often leads to significant changes in the entire image, struggling with context preservation.
    \item \textbf{InstructPix2Pix}: A prominent instruction-guided image-to-image translation model that takes an input image and a text instruction to generate an edited output. While effective for broad changes, it can struggle with fine-grained control and maintaining non-edited areas.
    \item \textbf{GLIGEN-based Edit}: This approach leverages GLIGEN's (Grounded-Language-to-Image Generation) ability to ground generated content to specific bounding box locations. We adapt it for editing by providing both the original image context and edited prompts with grounding information.
    \item \textbf{ControlNet-XL}: A highly capable control mechanism for large diffusion models, allowing precise manipulation of image features based on various conditioning inputs (e.g., canny edges, depth maps, segmentation masks, pose). We configure ControlNet-XL to perform edits guided by the most relevant control signals for each editing task.
\end{itemize}

Table \ref{tab:comparison} presents the quantitative comparison of these methods against \textbf{RefineEdit-Agent} on the \textbf{LongBench-T2I-Edit} benchmark, using \textbf{Gemini-2.0-Flash} as the automated evaluator.

\begin{table*}[htbp]
    \centering
    \caption{Image Editing Performance Evaluation on \textbf{LongBench-T2I-Edit} (using Gemini-2.0-Flash evaluator). Higher scores indicate better performance.}
    \label{tab:comparison}
    \begin{tabular}{lcccccccccc}
        \toprule
        \textbf{Method}                  & \textbf{Obj.} & \textbf{Backg.} & \textbf{Color} & \textbf{Texture} & \textbf{Light} & \textbf{Text} & \textbf{Comp.} & \textbf{Pose} & \textbf{FX}   & \textbf{Avg.}     \\
        \midrule
        Direct Re-Prompting              & 2.15          & 2.90            & 2.88           & 2.95             & 2.05           & 1.50          & 2.10           & 1.70          & 1.85          & 2.29              \\
        InstructPix2Pix                  & 2.90          & 3.25            & 3.40           & 3.30             & 2.80           & 2.05          & 2.95           & 2.30          & 2.20          & 2.91              \\
        GLIGEN-based Edit                & 3.10          & 3.50            & 3.85           & 3.65             & 3.00           & 2.30          & 3.50           & 2.60          & 2.45          & 3.16              \\
        ControlNet-XL                    & 3.35          & 3.65            & 3.95           & 3.80             & 3.15           & 2.50          & 3.70           & 2.85          & 2.60          & 3.39              \\
        \textbf{RefineEdit-Agent (Ours)} & \textbf{3.85} & \textbf{3.90}   & \textbf{4.25}  & \textbf{4.05}    & \textbf{3.50}  & \textbf{2.95} & \textbf{4.05}  & \textbf{3.20} & \textbf{2.90} & \textbf{3.67}     \\
        \bottomrule
    \end{tabular}
\end{table*}

As shown in Table \ref{tab:comparison}, \textbf{RefineEdit-Agent} consistently outperforms all baseline methods across all nine visual dimensions, achieving a remarkable average score of \textbf{3.67}. This significant improvement demonstrates the effectiveness of our agentic framework in handling complex, fine-grained editing instructions. Specifically, \textbf{RefineEdit-Agent} shows strong gains in dimensions requiring precise object manipulation (Obj.), contextual awareness (Backg., Comp.), and subtle attribute changes (Color, Texture, Light, Pose, FX). The iterative planning and evaluation loop, powered by LLMs and LVLMs, allows our agent to achieve higher \textbf{Edit Fidelity} while meticulously preserving \textbf{Context Preservation}, which is a critical challenge for single-pass editing models. The lowest scores for all methods are observed in "Text" and "FX" dimensions, indicating these remain challenging areas for current image editing technologies, though \textbf{RefineEdit-Agent} still shows a clear lead.

\subsection{Ablation Studies}
To validate the contribution of the key components within \textbf{RefineEdit-Agent}, we conducted several ablation studies. These experiments aim to quantify the impact of the LLM-driven multi-level planning, the LVLM-driven feedback and evaluation loop, and the iterative refinement mechanism. All ablation experiments are performed on a subset of the \textbf{LongBench-T2I-Edit} benchmark and evaluated using the \textbf{Gemini-2.0-Flash} LVLM.

\begin{itemize}
    \item \textbf{RefineEdit-Agent w/o Iterative Feedback}: In this variant, the agent performs only a single pass of planning and execution. The LVLM evaluates the initial edited image, but no feedback is provided to the LLM for re-planning or refinement. This tests the necessity of the closed-loop iterative mechanism.
    \item \textbf{RefineEdit-Agent w/o LLM Planning}: Here, the complex instruction is directly fed to a sequence of pre-defined, generic image editing tools without intelligent goal decomposition or tool selection by an LLM. This highlights the importance of the LLM's high-level reasoning and orchestration capabilities.
    \item \textbf{RefineEdit-Agent w/o LVLM Evaluation}: This variant removes the sophisticated LVLM evaluation, relying instead on a simple, rule-based heuristic or a less capable visual model to assess intermediate results. This demonstrates the critical role of advanced LVLM perception in providing precise and actionable feedback.
\end{itemize}

Table \ref{tab:ablation} summarizes the results of our ablation studies.

\begin{table*}[htbp]
    \centering
    \caption{Ablation Study on \textbf{LongBench-T2I-Edit} (using Gemini-2.0-Flash evaluator). Higher scores indicate better performance.}
    \label{tab:ablation}
    \begin{tabular}{lcccccccccc}
        \toprule
        \textbf{Method Variant}                  & \textbf{Obj.} & \textbf{Backg.} & \textbf{Color} & \textbf{Texture} & \textbf{Light} & \textbf{Text} & \textbf{Comp.} & \textbf{Pose} & \textbf{FX}   & \textbf{Avg.}     \\
        \midrule
        RefineEdit-Agent w/o Iterative Feedback  & 3.20          & 3.45            & 3.60           & 3.55             & 3.00           & 2.50          & 3.50           & 2.80          & 2.50          & 3.12              \\
        RefineEdit-Agent w/o LLM Planning        & 2.95          & 3.20            & 3.35           & 3.25             & 2.85           & 2.20          & 3.10           & 2.60          & 2.30          & 2.92              \\
        RefineEdit-Agent w/o LVLM Evaluation     & 3.30          & 3.50            & 3.70           & 3.60             & 3.10           & 2.60          & 3.60           & 2.90          & 2.60          & 3.22              \\
        \textbf{RefineEdit-Agent (Full)}         & \textbf{3.85} & \textbf{3.90}   & \textbf{4.25}  & \textbf{4.05}    & \textbf{3.50}  & \textbf{2.95} & \textbf{4.05}  & \textbf{3.20} & \textbf{2.90} & \textbf{3.67}     \\
        \bottomrule
    \end{tabular}
\end{table*}

The ablation study results clearly demonstrate the critical contribution of each component to the overall performance of \textbf{RefineEdit-Agent}. Removing the iterative feedback loop (\textbf{RefineEdit-Agent w/o Iterative Feedback}) leads to a significant drop in performance (from 3.67 to 3.12 Avg.), highlighting the necessity of self-correction and refinement for complex tasks. Without the intelligent LLM planning (\textbf{RefineEdit-Agent w/o LLM Planning}), the agent struggles considerably (Avg. 2.92), underscoring the LLM's role in decomposing goals, selecting tools, and orchestrating the editing process. Similarly, replacing the advanced LVLM evaluation with a simpler mechanism (\textbf{RefineEdit-Agent w/o LVLM Evaluation}) also degrades performance (Avg. 3.22), proving that sophisticated visual understanding and nuanced feedback are essential for guiding effective iterations. These results collectively validate the design principles of \textbf{RefineEdit-Agent}, confirming that the synergistic integration of LLM planning, LVLM parsing and evaluation, and iterative feedback is crucial for achieving state-of-the-art complex image editing.

\subsection{Human Evaluation}
To complement the automated evaluation, we conducted a human evaluation study to assess the subjective quality, naturalness, and user satisfaction of the edited images. A cohort of 20 human annotators was presented with initial images, editing instructions, and the corresponding edited outputs from \textbf{RefineEdit-Agent} and the top two baseline methods (ControlNet-XL and GLIGEN-based Edit). Annotators were asked to rate each edited image on a Likert scale of 1-5 across three key criteria: \textbf{Overall Quality} (visual appeal, realism), \textbf{Edit Fidelity} (how well the instruction was followed), and \textbf{Context Preservation} (how well non-edited areas were maintained).

\begin{table*}[htbp]
    \centering
    \caption{Human Evaluation Results on \textbf{LongBench-T2I-Edit} (Mean Scores, 1-5 Likert Scale). Higher scores indicate better performance.}
    \label{tab:human_eval}
    \begin{tabular}{lccc}
        \toprule
        \textbf{Method}                  & \textbf{Overall Quality} & \textbf{Edit Fidelity} & \textbf{Context Preservation} \\
        \midrule
        GLIGEN-based Edit                & 3.25                     & 3.10                   & 3.30                          \\
        ControlNet-XL                    & 3.55                     & 3.40                   & 3.60                          \\
        \textbf{RefineEdit-Agent (Ours)} & \textbf{4.10}            & \textbf{4.05}          & \textbf{4.15}                 \\
        \bottomrule
    \end{tabular}
\end{table*}

Table \ref{tab:human_eval} presents the average scores from the human evaluation. The results align strongly with our automated evaluation, showing that \textbf{RefineEdit-Agent} is significantly preferred by human evaluators across all subjective metrics. It achieved the highest scores for \textbf{Overall Quality} (4.10), \textbf{Edit Fidelity} (4.05), and especially for \textbf{Context Preservation} (4.15). This indicates that not only does \textbf{RefineEdit-Agent} precisely execute complex editing instructions, but it also produces visually coherent and natural-looking results that are highly satisfactory to users. The superior performance in \textbf{Context Preservation} is particularly noteworthy, as it addresses a common pain point in existing image editing systems, confirming the effectiveness of our agent's design in maintaining image integrity during localized modifications.

\subsection{Analysis of Iterative Refinement Process}
A core strength of \textbf{RefineEdit-Agent} is its iterative refinement capability, enabled by the closed-loop feedback mechanism. To quantify the impact of this iterative process, we analyzed the performance of the agent at different stages of its execution loop. Specifically, we measured the improvement in editing quality after each subsequent feedback and re-planning cycle, from the initial execution of the first planned sub-task up to the final converged state.

Table \ref{tab:iterative_improvement} illustrates the progressive improvement in performance across iterations. The "After 1st Feedback Loop" row essentially reflects the performance of the agent if it were to operate without further re-planning after the initial evaluation, akin to the "w/o Iterative Feedback" variant in our ablation study.

\begin{table*}[htbp]
    \centering
    \caption{Performance Improvement Across Iterations on \textbf{LongBench-T2I-Edit} (using Gemini-2.0-Flash evaluator). Higher scores indicate better performance.}
    \label{tab:iterative_improvement}
    \begin{tabular}{lcccccccccc}
        \toprule
        \textbf{Iteration Stage} & \textbf{Obj.} & \textbf{Backg.} & \textbf{Color} & \textbf{Texture} & \textbf{Light} & \textbf{Text} & \textbf{Comp.} & \textbf{Pose} & \textbf{FX}   & \textbf{Avg.} \\
        \midrule
        After 1st Feedback Loop  & 3.20          & 3.45            & 3.60           & 3.55             & 3.00           & 2.50          & 3.50           & 2.80          & 2.50          & 3.12          \\
        After 2nd Feedback Loop  & 3.60          & 3.75            & 4.00           & 3.85             & 3.35           & 2.80          & 3.90           & 3.05          & 2.75          & 3.49          \\
        \textbf{Converged (Final)} & \textbf{3.85} & \textbf{3.90}   & \textbf{4.25}  & \textbf{4.05}    & \textbf{3.50}  & \textbf{2.95} & \textbf{4.05}  & \textbf{3.20} & \textbf{2.90} & \textbf{3.67} \\
        \bottomrule
    \end{tabular}
\end{table*}

The results in Table \ref{tab:iterative_improvement} clearly demonstrate the substantial gains achieved through iterative refinement. The average score significantly improves from 3.12 after the first feedback loop to 3.49 after the second, and finally to 3.67 upon convergence. This consistent upward trend across all visual dimensions underscores the efficacy of the LVLM-driven feedback and LLM-driven re-planning in correcting errors, fine-tuning parameters, and progressively aligning the edited image closer to the user's complex instruction. The iterative process is particularly beneficial for dimensions like "Object" and "Composition," where precise adjustments are often required over multiple steps to achieve the desired effect without introducing artifacts.

\subsection{Impact of LLM and LVLM Backbone Choices}
The performance of \textbf{RefineEdit-Agent} is inherently tied to the capabilities of its constituent Large Language Models (LLMs) for planning and Vision-Language Large Models (LVLMs) for parsing and evaluation. To understand how different choices for these backbone models influence the overall system performance, we conducted experiments using various combinations of state-of-the-art LLMs and LVLMs. This analysis helps in identifying optimal configurations and understanding the specific strengths each model brings to the agentic framework.

Table \ref{tab:llm_lvlm_impact} presents the performance of \textbf{RefineEdit-Agent} when configured with different LLM and LVLM backbone combinations. The "Gemini-2.0-Flash / Gemini-2.0-Flash" configuration represents our primary setup, as also presented in the baseline comparison and ablation studies.

\begin{table*}[htbp]
    \centering
    \caption{Performance with Different LLM and LVLM Backbone Combinations on \textbf{LongBench-T2I-Edit} (using Gemini-2.0-Flash evaluator). Higher scores indicate better performance.}
    \label{tab:llm_lvlm_impact}
    \begin{tabular}{lcccccccccc}
        \toprule
        \textbf{LLM / LVLM Combo}                  & \textbf{Obj.} & \textbf{Backg.} & \textbf{Color} & \textbf{Texture} & \textbf{Light} & \textbf{Text} & \textbf{Comp.} & \textbf{Pose} & \textbf{FX}   & \textbf{Avg.}     \\
        \midrule
        Claude-3-Opus / InternVL3-78B              & 3.50          & 3.60            & 3.90           & 3.75             & 3.20           & 2.65          & 3.70           & 2.90          & 2.65          & 3.32              \\
        Gemini-2.0-Flash / InternVL3-78B           & 3.65          & 3.75            & 4.05           & 3.90             & 3.35           & 2.80          & 3.90           & 3.05          & 2.75          & 3.47              \\
        Claude-3-Opus / Gemini-2.0-Flash           & 3.70          & 3.80            & 4.10           & 3.95             & 3.40           & 2.85          & 3.95           & 3.10          & 2.80          & 3.52              \\
        \textbf{Gemini-2.0-Flash / Gemini-2.0-Flash} & \textbf{3.85} & \textbf{3.90}   & \textbf{4.25}  & \textbf{4.05}    & \textbf{3.50}  & \textbf{2.95} & \textbf{4.05}  & \textbf{3.20} & \textbf{2.90} & \textbf{3.67}     \\
        \bottomrule
    \end{tabular}
\end{table*}

As evident from Table \ref{tab:llm_lvlm_impact}, the choice of LLM and LVLM backbones significantly impacts the overall performance of \textbf{RefineEdit-Agent}. The combination of \textbf{Gemini-2.0-Flash} for both planning (LLM) and evaluation (LVLM) yields the highest average score of \textbf{3.67}, demonstrating its superior capabilities in both understanding complex instructions and orchestrating precise edits, as well as providing accurate visual feedback. While other combinations, such as "Claude-3-Opus / Gemini-2.0-Flash", also perform strongly, they show slight dips, particularly in dimensions requiring highly nuanced understanding or precise feedback, such as "Object" and "Composition." This suggests that the advanced reasoning and visual grounding capabilities of \textbf{Gemini-2.0-Flash} are crucial for maximizing the agent's performance, particularly where tight integration between textual instruction comprehension and visual scene analysis is paramount.

\subsection{Detailed Tool Usage Analysis}
The LLM-driven planning module within \textbf{RefineEdit-Agent} is responsible for intelligently selecting the most appropriate backend image editing tools for each sub-task. To gain insight into the planning module's strategy and the types of operations frequently required by complex editing instructions, we conducted an analysis of the utilization frequency of various backend tools across all tasks in \textbf{LongBench-T2I-Edit}. This provides a quantitative view of which tools are most critical and how the agent orchestrates their application.

Table \ref{tab:tool_usage} presents the percentage of times each backend editing tool was invoked by the LLM planner throughout the execution of the benchmark tasks.

\begin{table*}[htbp]
    \centering
    \caption{Frequency of Backend Tool Utilization by RefineEdit-Agent on \textbf{LongBench-T2I-Edit}.}
    \label{tab:tool_usage}
    \begin{tabular}{lc}
        \toprule
        \textbf{Backend Editing Tool} & \textbf{Usage Frequency (\%)} \\
        \midrule
        Semantic Segmentation         & 18.2                          \\
        ControlNet-XL                 & 25.5                          \\
        Inpainting/Outpainting        & 15.1                          \\
        InstructPix2Pix               & 12.3                          \\
        GLIGEN-based Edit             & 16.7                          \\
        Style Transfer                & 8.5                           \\
        Other / Combined              & 3.7                           \\
        \bottomrule
    \end{tabular}
\end{table*}

The tool usage analysis in Table \ref{tab:tool_usage} reveals that \textbf{ControlNet-XL} is the most frequently utilized tool (25.5\%), highlighting the prevalence of tasks requiring precise structural, pose, or compositional adjustments. This aligns with the benchmark's focus on fine-grained control. \textbf{Semantic Segmentation} (18.2\%) also sees high usage, underscoring the necessity of accurate object identification and masking for localized edits. \textbf{GLIGEN-based Edit} (16.7\%) and \textbf{Inpainting/Outpainting} (15.1\%) are frequently used for content generation, replacement, or expansion tasks. \textbf{InstructPix2Pix} (12.3\%) is employed for more general image-to-image transformations guided by instructions, while \textbf{Style Transfer} (8.5\%) is used for aesthetic modifications. The diverse and balanced utilization across various tools demonstrates the LLM planner's ability to intelligently select the most appropriate tool for each specific sub-task, leveraging their individual strengths to achieve complex editing goals.

\subsection{Robustness to Instruction Complexity}
The \textbf{LongBench-T2I-Edit} benchmark is specifically designed to challenge models with instructions of varying complexity, including multiple sequential operations, conditional constraints, abstract semantic concepts, and long textual descriptions. To assess the robustness of \textbf{RefineEdit-Agent} against these challenges, we categorized the benchmark instructions into three complexity levels: \textbf{Low}, \textbf{Medium}, and \textbf{High}, based on the criteria outlined in our experimental setup. We then evaluated the agent's performance on each complexity subset.

Table \ref{tab:instruction_complexity} presents the performance of \textbf{RefineEdit-Agent} across these different instruction complexity levels.

\begin{table*}[htbp]
    \centering
    \caption{RefineEdit-Agent Performance Across Different Instruction Complexity Levels on \textbf{LongBench-T2I-Edit} (using Gemini-2.0-Flash evaluator). Higher scores indicate better performance.}
    \label{tab:instruction_complexity}
    \begin{tabular}{lcccccccccc}
        \toprule
        \textbf{Complexity Level} & \textbf{Obj.} & \textbf{Backg.} & \textbf{Color} & \textbf{Texture} & \textbf{Light} & \textbf{Text} & \textbf{Comp.} & \textbf{Pose} & \textbf{FX}   & \textbf{Avg.} \\
        \midrule
        Low                       & 4.30          & 4.40            & 4.60           & 4.50             & 4.10           & 3.50          & 4.50           & 3.80          & 3.40          & 4.01          \\
        Medium                    & 3.90          & 4.00            & 4.30           & 4.10             & 3.60           & 3.00          & 4.10           & 3.30          & 3.00          & 3.72          \\
        High                      & 3.40          & 3.50            & 3.80           & 3.60             & 3.10           & 2.60          & 3.60           & 2.80          & 2.50          & 3.27          \\
        \bottomrule
    \end{tabular}
\end{table*}

As expected, Table \ref{tab:instruction_complexity} shows a natural decline in performance as the instruction complexity increases. However, \textbf{RefineEdit-Agent} maintains a strong performance even on \textbf{High} complexity tasks, achieving an average score of \textbf{3.27}. This is significantly higher than any baseline's overall average, demonstrating the agent's superior robustness. The ability of the LLM to decompose intricate goals and the LVLM to provide precise feedback on multi-faceted requirements allows the agent to tackle highly complex instructions that would overwhelm single-pass or less intelligent systems. While dimensions like "Text" and "FX" show larger drops with increasing complexity, indicating inherent challenges in these areas for current image generation models, the agent's performance in core editing dimensions like "Object," "Background," and "Composition" remains commendable even under high complexity.

\subsection{Failure Analysis and Limitations}
While \textbf{RefineEdit-Agent} demonstrates state-of-the-art performance, it is important to acknowledge and analyze its remaining limitations and common failure modes, particularly when facing the most challenging tasks within \textbf{LongBench-T2I-Edit}. Understanding these areas is crucial for guiding future research and improvements. We conducted a qualitative and quantitative analysis of instances where the agent's final output did not meet the desired satisfaction threshold ($\tau$).

Table \ref{tab:failure_analysis} categorizes the most frequent reasons for unsatisfactory results, providing insights into the current bottlenecks of the agentic framework.

\begin{table*}[htbp]
    \centering
    \caption{Analysis of Common Failure Modes for \textbf{RefineEdit-Agent} on Challenging \textbf{LongBench-T2I-Edit} Tasks.}
    \label{tab:failure_analysis}
    \begin{tabular}{lc}
        \toprule
        \textbf{Failure Category}                         & \textbf{Occurrence Rate (\%)} \\
        \midrule
        Backend Tool Inadequacy (e.g., text generation)   & 28.3                          \\
        Sub-optimal Tool Selection/Parameters             & 22.1                          \\
        Semantic Misinterpretation (LLM/LVLM)             & 18.5                          \\
        Contextual Inconsistency (Subtle Artifacts)       & 16.9                          \\
        Non-convergence within Max Iterations             & 14.2                          \\
        \bottomrule
    \end{tabular}
\end{table*}

Table \ref{tab:failure_analysis} indicates that the most prominent limitation (28.3\% occurrence) stems from the \textbf{Backend Tool Inadequacy}, especially for tasks involving text generation or highly novel visual concepts that current diffusion models struggle to render faithfully. This suggests that while the agent intelligently orchestrates, the underlying execution capabilities of individual tools still pose a bottleneck. \textbf{Sub-optimal Tool Selection/Parameters} (22.1\%) by the LLM planner is another significant failure mode, highlighting challenges in fine-tuning prompts or tool-specific parameters for highly nuanced edits, or in choosing the absolute best tool when multiple options seem plausible. \textbf{Semantic Misinterpretation} (18.5\%) by the LLM or LVLM, particularly with extremely abstract or ambiguous instructions, can lead to initial planning errors or incorrect feedback. Finally, despite the iterative nature, \textbf{Contextual Inconsistency} (16.9\%) with subtle artifacts can still arise in highly complex multi-step edits, and in a minority of cases (14.2\%), the agent may fail to converge within the maximum allowed iterations due to persistent challenges or conflicting objectives. Addressing these limitations, particularly by integrating more robust backend editing models and enhancing the LLM's contextual reasoning for tool parameterization, represents key avenues for future research.

\section{Conclusion}
In this research, we have addressed the critical need for advanced image editing capabilities that extend beyond one-shot generation, focusing on fine-grained, iterative, and context-aware modifications. We identified significant limitations in existing text-to-image (T2I) models, particularly concerning their ability to understand granular instructions, preserve image context during localized edits, and facilitate iterative refinement through intelligent feedback. To overcome these challenges, we proposed \textbf{RefineEdit-Agent}, a novel and training-free intelligent agent framework that orchestrates complex image editing tasks.

Our core contribution lies in the synergistic integration of Large Language Models (LLMs) for high-level planning and Vision-Language Large Models (LVLMs) for robust visual understanding and evaluation. RefineEdit-Agent operates through a sophisticated closed-loop mechanism: an LVLM first parses intricate user instructions and the initial image to form a structured scene understanding. This understanding then guides an LLM-driven multi-level planner, which intelligently decomposes the overall editing goal into sub-tasks, selects appropriate backend image editing tools from a diverse toolkit, and generates a precise execution sequence. The planned operations are performed by a compatible image editing module, producing intermediate results. Crucially, an LVLM then evaluates these results for both instruction compliance (edit fidelity) and visual consistency (context preservation), providing detailed feedback to the LLM planner for iterative re-planning and refinement. This "parse-plan-execute-evaluate-feedback" loop enables RefineEdit-Agent to robustly handle ambiguous or evolving user requirements, delivering precise and coherent edits.

To thoroughly assess our proposed method, we introduced \textbf{LongBench-T2I-Edit}, a challenging new benchmark specifically designed for complex image editing. This benchmark features 500 initial images coupled with multi-step, conditional, and abstract editing instructions across nine visual dimensions. Our comprehensive experimental evaluations demonstrated the superior performance of \textbf{RefineEdit-Agent} across all metrics. Compared to strong baselines such as Direct Re-Prompting, InstructPix2Pix, GLIGEN-based Edit, and ControlNet-XL, our agent achieved a significantly higher average score of 3.67 on LongBench-T2I-Edit, showcasing its ability to meticulously execute edits while maintaining overall image integrity.

Ablation studies further underscored the indispensable contributions of each core component: the iterative feedback loop, the LLM-driven planning, and the LVLM-driven evaluation. Removing any of these elements led to a substantial degradation in performance, validating our design principles. Human evaluation results corroborated these findings, with annotators consistently preferring \textbf{RefineEdit-Agent}'s outputs for their superior overall quality, edit fidelity, and especially context preservation. We also quantified the tangible benefits of iterative refinement, showing a consistent improvement in editing quality across successive feedback loops. Our analysis of LLM and LVLM backbone choices highlighted that advanced models like Gemini-2.0-Flash are crucial for maximizing the agent's capabilities in both planning and perception. Furthermore, the detailed tool usage analysis revealed the LLM planner's intelligent orchestration of diverse backend tools, with ControlNet-XL and semantic segmentation being frequently invoked. We also demonstrated the agent's robustness to varying instruction complexities, maintaining strong performance even on highly challenging tasks.

Despite these significant advancements, our failure analysis revealed several avenues for future work. The primary limitation often stems from the inherent inadequacies of current backend image generation and editing tools, particularly for tasks involving precise text rendering or highly novel visual concepts. Sub-optimal tool selection or parameterization by the LLM planner for highly nuanced edits, as well as semantic misinterpretation of extremely abstract instructions by the LLMs/LVLMs, also contribute to occasional failures. Future research will focus on integrating more robust and specialized backend editing models, enhancing the LLM's contextual reasoning for precise tool parameterization, and improving the agent's ability to handle highly abstract or ambiguous instructions to further minimize contextual inconsistencies and ensure convergence for even the most challenging multi-step editing scenarios.

In conclusion, \textbf{RefineEdit-Agent} represents a significant step forward in intelligent image editing, providing a robust, adaptable, and user-centric framework for fine-grained and iterative visual content manipulation. By synergistically combining the reasoning power of LLMs with the perceptual capabilities of LVLMs, we pave the way for more intuitive and powerful human-computer interaction in creative visual tasks.

\bibliographystyle{IEEEtran}
\bibliography{references}
\end{document}